# Audience Engagement with Arabic Women's Social Empowerment and Wellbeing: A Decadal Corpus

**Wajdi Zaghouani**[1], **Mabrouka Bessghaier**[1], **MD. Rafiul Biswas**[2], **Shimaa Amer Ibrahim**[1]
[1] Northwestern University in Qatar
[2] Hamad bin Khalifa University, Qatar
{wajdi.zaghouani, mabrouka.bessghaier, shimaa.ibrahim}@northwestern.edu
mbiswas@hbku.edu.qa

**Abstract**

This paper presents the Arabic Women and Society Corpus, a ten-year collection of 252,487 public Arabic Facebook posts related to women's empowerment and social wellbeing. The corpus was collected from 51,660 pages across 77 countries between 2013 and 2024, resulting in more than 267 million user interactions. Each post includes engagement metrics such as shares, comments, and emotional reactions, providing a unique view of audience sentiment and social attention. The data were processed using an automated pipeline with language identification, normalization, and metadata cleaning to ensure reliability and reproducibility. The corpus enables large-scale analysis of gender discourse, social reform, and emotional engagement across Arabic dialects. It supports research in Arabic natural language processing, computational social science, and digital communication studies. The dataset and accompanying documentation will be released under request for research use.

## 1. Introduction

Women's participation in society and their empowerment have long been central to development agendas worldwide, including questions of gender equality, social justice, and human rights. In the Arab world, these discussions carry particular weight, reflecting complex interactions between cultural traditions, religious interpretations, political structures, and evolving social norms. Understanding how such issues are framed, debated, and received in public discourse is essential for evaluating both the challenges and progress of women's empowerment across the region. In fact, the concept of empowerment is a complex and cross-disciplinary process. It includes various subjects such as sociology, psychology, and economics, as well as issues such as education, healthcare, governance, and economic policy (Al-Maimani, 2021).

Social media platforms have emerged as crucial spaces for these conversations, serving not merely as communication tools but as public forums where sensitive topics are discussed, opinions are voiced, and social change is negotiated. Facebook, with its extensive reach across the Middle East and North Africa, provides unique opportunities to capture how diverse communities discuss women's roles in society and how audiences respond emotionally to such content. The patterns of engagement (i.e., likes, comments, shares, and emotional reactions such as Love, Angry, or Sad) offer valuable insights into broader social attitudes and cultural dynamics.

While several datasets exist for Arabic sentiment and emotion analysis, most remain limited in scale or focused on textual sentiment classification. Few resources capture the nuanced emotional dimensions of audience engagement through naturally occurring signals like Facebook reactions, and even fewer address sensitive themes such as women's empowerment with the breadth and depth necessary for comprehensive analysis. The scarcity of large-scale, longitudinal datasets dedicated to gender discourse in Arabic contexts represents a significant gap in computational social science resources, particularly given the central role these topics play in contemporary Arab societies.

This paper introduces the Women's Social Empowerment and Wellbeing corpus, a comprehensive dataset of public Facebook posts spanning over a decade (2013-2024). The corpus contains 252,487 posts from 51,660 unique pages across 77 countries, focusing on Arabic social media discourse about women and society. Collected using targeted Arabic keywords related to women's social participation, empowerment, and gender issues, the dataset includes rich metadata such as page attributes, detailed engagement metrics, and six types of emotional reactions. With the discontinuation of CrowdTangle in 2024, this corpus represents a unique historical archive of Arabic Facebook discourse that cannot be replicated.

## 2. Related Work

Research at the intersection of women's empowerment, digital discourse, and computational social science has expanded considerably. Yet significant gaps remain in the availability of large-scale, culturally grounded datasets that capture both content and patterns of emotional engagement.

A growing body of research has explored how Arabic women's rights and empowerment are represented in digital spaces. For example, Hurley (2021) analyzed Instagram self-presentations by Arab women, demonstrating how empowerment is negotiated through visibility, cultural norms, and technological affordances. While such qualitative studies provide important interpretive insights, they are typically limited in scale and temporal scope. More recent quantitative approaches have begun to address these limitations. In an eight-and-a-half-year analysis of Facebook posts from Egypt, Tunisia, and Jordan using CrowdTangle, El Baradei et al. (2025) applied the Dragonfly Advocacy Model to examine government-led communication on women's issues. They found that women's empowerment, eliminating violence, and ending harmful practices were the most frequently promoted themes by government pages, whereas topics like reproductive health and digital inclusion received less attention. However, this study was restricted to official advocacy pages and a narrower thematic scope. Moreover, Al-Maimani (2021) conducted thematic analysis of Arab women's empowerment via social media, finding that while digital platforms afford increased visibility, confidence, and opportunities for collective action, persistent cultural, legal, and familial constraints limit the translation of online gains into sustained offline empowerment. In addition, research by Gangwani et al. (2021) on social media usage and female empowerment in Saudi Arabia highlighted low participation rates among Saudi women on social platforms, despite the potential for empowerment. Our corpus complements and extends this work by offering women's empowerment discourse at scale, with 252K Facebook posts over a decade from thousands of pages in 77 countries. Covering themes such as violence and harassment, economic empowerment, and political participation, the dataset pairs rich thematic coverage with detailed audience reaction metrics, revealing how empowerment narratives resonate across diverse Arabic-speaking publics. Its scale and transparency enable replication and a wide range of follow-up research.

Computational approaches to understanding discourse have advanced rapidly, particularly in Arabic sentiment analysis and emotion classification (Aladeemy et al., 2024; Al Katat et al., 2024). Several datasets have been developed to support Arabic Natural Language Processing (NLP) research. Foundational resources such as ASTD (Nabil et al., 2015), ASAD (Alharbi et al., 2020), and ATSAD (Kwaik et al., 2020) have established key benchmarks for sentiment analysis. Additional datasets have been introduced for related tasks including abusive and hate speech detection (Mubarak et al., 2017; Charfi et al., 2024c,a), irony detection (Charfi et al., 2024d), and stance classification (Charfi et al., 2024b). More recent work has explored affective phenomena such as hope and hate emotion detection (Zaghouani et al., 2025). In parallel, methodological advances have improved sentiment and emotion modeling in Arabic. Transformer-based models have achieved state-of-the-art performance across multiple dialectal sentiment classification tasks (Alosaimi et al., 2024; Ibrahim et al., 2025). Emotion classification has also emerged as a complementary field that captures discrete affective states such as joy, anger, sadness, and fear (Alqahtani and Alothaim, 2022; Biswas et al., 2025). Despite these advances, most available datasets and approaches remain focused on analyzing sentiment, emotion, or intent expressed within the text itself. In many cases, datasets provide only coarse sentiment polarity labels and rarely incorporate signals of audience engagement or reaction-level information. For example, Duwairi and Qarqaz (2017) compiled an Arabic Facebook comment dataset for polarity classification without including reaction statistics.

Little attention has been given to audience-expressed emotions: how readers or viewers emotionally respond to the content they consume. These reactions reflect collective affective responses and the broader social resonance of discourse. While sentiment classification captures how individuals express emotion, audience engagement reveals how communities collectively respond. Our work addresses this complementary and underexplored dimension by focusing on audience engagement as a natural signal of emotional response. We analyze the distribution and intensity of audience interactions (e.g., likes, comments, shares) and their emotional tenor as reflected in Facebook's six reaction types (Love, Angry, Sad, Haha, Care, Wow). These naturally occurring engagement signals provide large-scale, real-world evidence of how Arabic-speaking audiences emotionally respond to gender discourse, uncovering shared moods and sparks of solidarity or dissent that shape public conversations around women's empowerment. To our knowledge, this is the first study to systematically capture reaction distributions in Arabic discourse on women's empowerment, bridging content analysis and affective response. In sum, prior work has advanced under-standing of women's empowerment online and developed sentiment resources; however, our dataset is thematically focused, linguistically and geographically diverse, and enriched with audience reaction signals, providing a foundation for future interdisciplinary research across gender studies, computational social science, and digital humanities.

## 3. Methodology: Dataset Creation

The dataset focuses on discourse around women's participation in society, including their workforce engagement and economic empowerment. It was specifically constructed to capture public discussions and patterns of audience engagement on Facebook. Posts were collected over a ten-year period from thousands of pages across diverse countries, ensuring both temporal depth and geographical breadth. This section describes the data collection process and presents the structure of the resulting dataset.

### 3.1. Data Collection

The dataset was collected from public Facebook[1] pages using the CrowdTangle API,[2] which makes the present dataset a historical archive of engagement data on the topic of women's empowerment. Only publicly available content (pages and selected public groups) was accessible through CrowdTangle.

The collection focused on posts containing a set of Arabic keywords related to women's social participation, empowerment, and gender issues. Keywords were drawn from three thematic clusters: (1) core empowerment terms (e.g., [illegible] [women's empowerment], [illegible] [women's rights], [illegible] [working woman]); (2) social and legal issues (e.g., [illegible] [violence against women], [illegible] [sexual harassment], [illegible] [family laws]); and (3) public life and identity (e.g., [illegible] [women in politics], [illegible] [International Women's Day], [illegible] [Hijab]). The full keyword list will be included in the dataset documentation released alongside the corpus. This approach captured the full spectrum of discussions around women's roles in society, from workplace participation to social advocacy and political engagement. Pages were not pre-selected; rather, all public pages that published at least one post matching the keyword criteria during the collection window were included. This surface-driven approach ensured diversity across page types (news organizations, advocacy groups, government entities, educational institutions, and individual activists) and minimized selection bias.

**Temporal Coverage:** The collection period spans from 19 November 2013 to 09 July 2024, capturing a full decade of significant social change in the Arab world, including post-Arab Spring developments, various reform initiatives, and evolving discussions around women's rights.

**Page Diversity:** The collection included diverse page types (e.g., news organizations, advocacy groups, government entities, educational institutions, and individual activists) to ensure representation of multiple perspectives and discourse styles.

### 3.2. Data Structure

The dataset contains a total of 41 variables describing page attributes, post content, audience size, engagement, sponsorship, and performance. These variables can be grouped into 11 categories, as detailed in Table 1.

At the **Page-level** category, the dataset captures identifying information such as *Page Name*, *User Name*, *Facebook ID*, *Page Category*, *Page Admin Top Country*, *Page Description*, and the *Page Created* date. **Page audience size at posting** is represented by the number of *Likes at Posting* and *Followers at Posting*. The dataset also includes **Post timing** metadata, covering three timestamp fields (*Post Created*, *Post Created Date*, and *Post Created Time*) and the declared **Post type** (*Type*, e.g., photo, video, link, or status).

Engagement metrics are represented at two levels: (a) **Engagement totals**, which aggregate counts of *Total Interactions*, *Likes*, *Comments*, and *Shares*; and (b) **Emotional reactions**, which include the distribution of reactions (*Love*, *Wow*, *Haha*, *Sad*, *Angry*, *Care*). For **Video-related** posts, additional fields capture ownership and sharing status (*Video Share Status*, *Is Video Owner?*), as well as exposure and duration metrics (*Post Views*, *Total Views*, *Total Views For All Crossposts*, *Video Length*).

The **Links and URLs** category includes the *URL* of the post, the originally shared *Link* (if any), and its resolved *Final Link* after redirects. The **Content and text fields** category captures textual information, including the post text in the *Message* column (status or caption), OCR-extracted content in the *Image Text* column, the link headline in *Link Text*, and the link preview snippet in *Description*. **Sponsorship** information is represented by *Sponsor Id*, *Sponsor Name*, and *Sponsor Category*, allowing the identification of branded or paid content. Finally, **Performance** indicators include the weighted *Total Interactions (weighted)* score and the *Overperforming Score*, which benchmarks each post's engagement against historical performance from the same page or category.
For reproducibility, each record also contains its associated URL.

[1] https://www.facebook.com

[2] The CrowdTangle API is a public insights tool from Meta designed to track and analyze public Facebook content. As of August 14, 2024, CrowdTangle has been discontinued and replaced by the Meta Content Library and Content Library API.

Table 1: Overview of the dataset structure with 41 columns grouped by category

| Aspect / Category | Description | Columns |
|---|---|---|
| Page-level (7) | Attributes of the source page | Page Name; User Name; Facebook Id; Page Category; Page Admin Top Country; Page Description; Page Created |
| Page audience at posting (2) | Page size at the time of posting | Likes at Posting; Followers at Posting |
| Post timing (3) | Timestamp fields for the post | Post Created; Post Created Date; Post Created Time |
| Post type (1) | Declared post type | Type (e.g. Video, Status, Photo, Link) |
| Engagement totals (4) | Aggregate engagement counts | Total Interactions; Likes; Comments; Shares |
| Emotional reactions (6) | Distribution of reactions | Love; Wow; Haha; Sad; Angry; Care |
| Video-related (6) | Ownership, exposure, and duration attributes of video posts | Video Share Status; Is Video Owner?; Post Views; Total Views; Total Views For All Crossposts; Video Length |
| Links & URLs (3) | Post and external link references | URL; Link; Final Link |
| Content & text fields (4) | Textual information | Message (post text or status); Image Text; Link Text; Description |
| Sponsorship (3) | Sponsor identifiers and taxonomy | Sponsor Id; Sponsor Name; Sponsor Category |
| Performance (2) | Derived performance indicators | Total Interactions (weighted); Overperforming Score |

## 4. Dataset Description and Statistics

This section provides a statistical overview of the dataset, highlighting its scale, engagement distribution, emotional reactions, content categories, and geographical reach.

### 4.1. Overall Statistics

The Arabic Women & Society corpus represents one of the largest collections of Arabic social media content focused on gender and empowerment topics. Spanning a full decade from 2014 to 2024, the corpus comprises **252,487** posts published by **51,660 unique public pages** across **77 countries**. The temporal span of ten years captures significant social and political developments across the Arab world, including the post-Arab Spring years, various social reform initiatives, and evolving discussions around women's rights and participation. The diversity of contributing pages, ranging from news outlets and advocacy organizations to individual activists and community groups, ensures representation of multiple perspectives and discourse styles.

### 4.2. Geographical Distribution and Cultural Diversity

The posts represent a diverse geographical distribution, with the top 10 contributing countries accounting for nearly 77% of all posts (see Figure 1): **Egypt** (75,676 posts), **Tunisia** (18,132 posts), **Iraq** (17,932 posts), **Algeria** (16,257 posts), **Syria** (14,344 posts), **Morocco** (13,433 posts), **Palestine** (13,316 posts), **Jordan** (10,646 posts), **Turkey** (7,276 posts), and **Libya** (6,560 posts). This highlights the dataset's strong representation of North African and Levantine countries. The remaining 67 countries collectively contribute approximately 23% of all posts; these include Gulf states such as Saudi Arabia, the United Arab Emirates, Kuwait, and Qatar, as well as diaspora communities in European countries (e.g., France, Germany, and Sweden), North American countries (e.g., the United States and Canada), and additional Arabic-speaking communities globally. A full country-level breakdown is provided in the dataset documentation. This geographical diversity ensures that the corpus captures a wide range of Arabic dialects, cultural contexts, and socio-political environments, providing a rich foundation for cross-dialectal and cross-cultural analysis of women's discourse in Arabic social media.

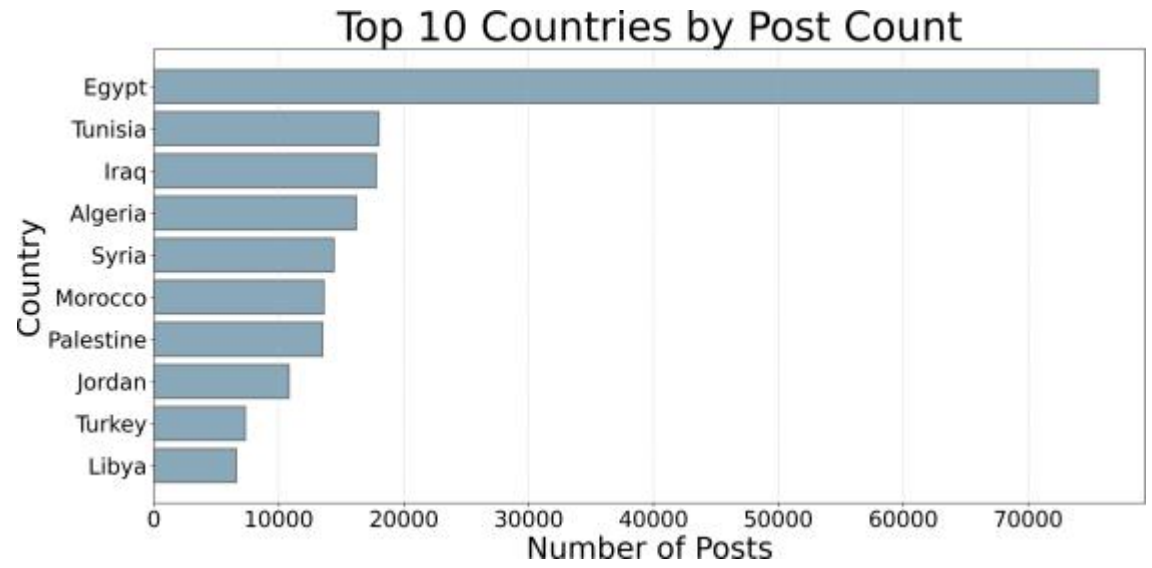

Figure 1: Distribution of posts by the top 10 countries of page administrators

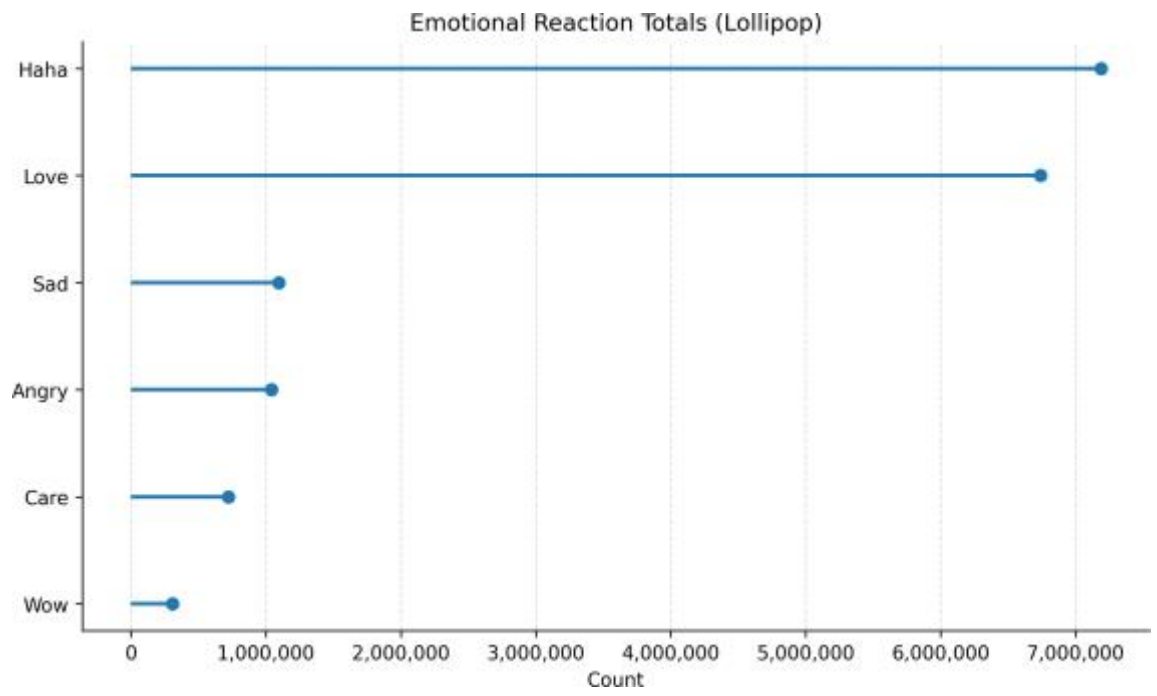

Figure 2: Distribution of total emotional reactions

### 4.3. Dialectal Composition

Given the geographical spread of contributing pages, the corpus naturally contains a variety of Arabic dialects alongside Modern Standard Arabic (MSA). The top contributing countries suggest that Egyptian Arabic, Maghrebi varieties (Moroccan, Algerian, and Tunisian Darija), and Levantine dialects (Syrian, Palestinian, and Jordanian) are prominently represented, while Gulf Arabic and Iraqi dialects are also present. Language identification was applied during preprocessing (Section 5.1) to flag non-Arabic tokens; however, dialectal variation is treated as a feature rather than noise, since cross-dialectal coverage is one of the corpus's primary strengths. Researchers who require dialect-level stratification are encouraged to apply dialect identification tools such as AIDA2 (**?**) or ALDi (**?**) to the released data. We include dialect identification as a recommended preprocessing step in the accompanying documentation.

### 4.4. Content Composition and Media Types

The corpus reveals clear preferences in content formats for discussing women's empowerment. Photo posts dominate with 194,806 entries (77.15%), indicating the visual nature of empowerment discourse. Link posts account for 32,615 entries (12.92%), showing significant sharing of external content. Native videos comprise 13,257 posts (5.25%), representing multimedia storytelling. Status updates make up 7,251 posts (2.87%), capturing direct commentary and opinions. Live videos total 2,552 posts (1.01%), indicating real-time engagement with audiences. YouTube shares number 1,644 (0.65%), showing cross-platform content integration. The distribution clearly shows that visual and multimedia content drive audience engagement on women's empowerment topics.

### 4.5. Engagement Distribution

Engagement across the dataset is exceptionally high, underscoring the sustained public interest in women's empowerment across Arabic social media. In total, the corpus records more than 57 million Likes, 12.9 million Comments, and 7.2 million Shares. Emotional reactions add another important layer, with nearly 6.7 million Love, 7.2 million Haha, 1.1 million Sad, 0.7 million Care, 1 million Angry, and 0.3 million Wow responses, as illustrated in Figure 2.

These figures reveal a complex affective landscape. The *Love* and *Haha* reactions together account for more than **80% of all emotional responses**, making them by far the most frequent. It is important, however, to acknowledge that Facebook reactions carry inherent interpretive ambiguity. *Love* reactions most plausibly signal approval, empathy, and solidarity with empowerment narratives, but they may also reflect engagement with ironic or satirical framings of the same issues. *Haha* reactions are similarly multivalent: they may indicate humor or lighthearted content, but they can equally express sarcasm, ridicule, or dismissal directed at empowerment arguments. *Angry* reactions are comparably ambiguous, as they could reflect outrage at injustice described in a post but could equally express opposition to its feminist message. These ambiguities cannot be fully resolved without post-level annotation, and we caution against interpreting reaction counts as direct proxies for specific stances. The reaction distributions reported here are best understood as aggregate indicators of affective salience rather than as unambiguous sentiment labels.

Negative and empathetic emotions appear less frequently but remain socially significant. Sad and Angry reactions together represent roughly 15% of total reactions, typically clustering around posts addressing violence, discrimination, or systemic injustice. Meanwhile, Care and Wow reactions, though comparatively rare, express compassion and admiration, reflecting moments of moral con-cern or inspiration.

When examining dominant reactions per post, the dataset shows that the majority of posts elicit overwhelmingly positive audience sentiment: 191,103 posts are dominated by Love reactions, compared to 37,162 Haha, 9,807 Sad, 7,337 Angry, 4,419 Care, and 2,659 Wow. This pattern is consistent with Arabic Facebook discourse on women's empowerment being primarily characterized by affirmation and support, while humor and emotional critique remain meaningful modes of public engagement.

Together, these engagement dynamics illustrate a nuanced spectrum of audience emotions, revealing how expressions of solidarity, irony, and empathy coexist within Arabic social media discourse on women's empowerment.

### 4.6. Data Availability

The Arabic Women and Society Corpus, together with its complete documentation, annotation schema, and preprocessing scripts, will be released under request for research use[3]. To clarify the release scope: the distributed version will include anonymized post texts, aggregated engagement metrics, country-level identifiers, and the full keyword list used for collection. Access is subject to a data-sharing agreement requiring (a) noncommercial academic use, (b) institutional ethics or IRB approval (or equivalent waiver), and (c) a prohibition on re-identification or de-anonymization of individuals. This approach balances reproducibility with responsible data governance, consistent with the ethical considerations discussed in Section 9. To support reproducibility, Python scripts and a detailed data dictionary describing all 41 variables will also be made available under the same agreement. All resources will be distributed under a Creative Commons Attribution NonCommercial ShareAlike 4.0 (CC BY-NC-SA 4.0) License to the extent permitted by applicable platform terms; researchers are advised to consult Meta's current terms of service regarding reuse of derived social-media data. Access will be provided through a permanent repository, linked in the final publication.

## 5. Use Case Analysis: Data-Driven Topic Discovery and Emotional Engagement

To demonstrate the analytical capabilities of our dataset, we present a comprehensive case study examining how different aspects of women's empowerment discourse generate distinct patterns of emotional engagement on Arabic social media.

[3] https://tinyurl.com/4ke5jwyw

Using an automated Arabic-optimized BERTopic pipeline (Grootendorst, 2022), we apply a data-driven approach to identify dominant discussion themes and quantify audience reactions. The pipeline integrates lexical frequency modeling, dimensionality reduction, and density-based clustering to produce interpretable, emotion-aware topic structures. This enables the exploration of how diverse women's narratives resonate emotionally across the Arabic-speaking online sphere.

### 5.1. Methodology

Our analysis pipeline consists of four main stages.

#### 5.1.1. Text Aggregation and Normalization

All available textual fields per record (including Message, Description, Link Text, and Image Text) were concatenated into a single textual unit after removing empty or duplicated entries. The resulting corpus was normalized through a tailored Arabic preprocessing pipeline that included diacritic and punctuation removal, orthographic normalization, and elimination of common Arabic stop words. Because the corpus spans multiple dialects, including heavily divergent Maghrebi varieties that blend Arabic with French or Berber lexical items, stop word removal was limited to a curated MSA list to avoid inadvertently stripping content-bearing dialectal tokens. A language identification step using `fastText` (Joulin et al., 2016) was applied to flag and quarantine non-Arabic records; these were excluded from topic modeling but retained in the released dataset for completeness.

#### 5.1.2. Topic Discovery and Representation

Topic discovery was conducted using the BERTopic framework (Grootendorst, 2022), adapted for Arabic text. To emphasize interpretability and ensure robustness across dialects (where contextual embeddings pre-trained on MSA may underperform), the model used lexical frequency information rather than contextual embeddings. Each document was represented by TF-IDF weighting based on the 4,000 most frequent Arabic unigrams and bigrams. The resulting high-dimensional feature space was reduced to 100 latent components through truncated singular value decomposition (SVD). Dimensionality reduction via UMAP (Ghojogh et al., 2021) captured local and global structure in a five-dimensional embedding, while HDBSCAN (McInnes et al., 2017) performed unsupervised clustering to identify semantically coherent groups. We constrained the maximum number of final topics to 20, a value chosen through iterative experimentation: preliminary runs with unconstrained HDBSCAN produced a large number of fine-grained micro-clusters, many of which were semantically redundant. Reducing to a maximum of 20 yielded general yet distinctive discourse themes whose coherence was more interpretable for human validation. Future work could apply dynamic topic modeling or finer granularity to recover subthemes within each cluster.

#### 5.1.3. Topic Assignment and Labeling

Each post was assigned to its most representative topic cluster, while unclustered posts were labeled as "Noise". The overall topic inventory, including topic size, label, and representative terms, was also summarized. This setup facilitates both quantitative exploration and qualitative inspection of topical meaning.

#### 5.1.4. Emotional Engagement Quantification

To capture audience sentiment and affective response, we incorporated post-level reaction data. To assess public sentiment toward each topic, we aggregated Facebook reactions (*Love*, *Haha*, *Wow*, *Sad*, *Angry*, *Care*) for all posts assigned to a given topic. Table 2 summarizes the results for the five most frequent topics.

### 5.2. Results

The BERTopic model automatically identified up to twenty coherent thematic clusters (excluding noise), reflecting the diversity of online discussions surrounding women's empowerment. Each topic was characterized by its top-ranked Arabic keywords, and coherence was verified qualitatively by examining representative posts sampled per cluster. Table 2 shows clear, topic-specific affective profiles. Overall, *Love* and *Haha* account for the bulk of reactions across themes, while negative and empathetic signals (*Sad*, *Angry*, *Care*) vary by topic. Representative topics included:

- **Violence and Discrimination:** includes discourse on gender-based violence, harassment, and legal protection (*e.g.,* ·ỏl_ll .)..,· ·-Jl [violence against women], f··›J l v''·_.,.··Jl [sexual harassment], l· l [rape]); this category dominated online engagement and triggered high *Sad* and *Care* reactions, reflecting collective empathy and outrage.
- **Religion and Morality:** clusters centered on religious framing and theological debates (*e.g.,* l›J l [Hijab], _../· l [Al-Azhar], ,l·· '' l [religious edict], l.l··Jl [Niqab]); often polarizing audiences between conservative and reformist views. Such topics showed elevated *Angry* reactions (10.4%), which may reflect both moral concern about perceived violations of religious norms and frustration directed at restrictive interpretations, underscoring the interpretive ambiguity of this reaction type.
- **Policy, Law, and Institutions:** governmental and organizational narratives around empowerment (*e.g.,* 0·l_..lJ f,. lJl .._l.,.. _ll [National Council for Women], 0· ,· l.l·Jl [law], fJ--.)Jl _r·_ll [international conference]), focusing on official programs and legal frameworks; these posts attracted steady but less emotionally charged engagement, dominated by *Love* reactions.
- **Gender Roles and Social Norms:** debates and commentary around interpersonal dynamics, marriage, and stereotypes (*e.g.,* J'''. _Jl [man], l'''. --_·Jl .)-,· [polygamy], Jl'''. _Jl .)· ,·,. [men are attracted]). This cluster is distinguished from "Cultural Identity and Celebration" by its focus on contested and prescriptive gender norms, everyday relational dynamics, and satirical commentary on social expectations, rather than on symbolic achievements or national identity. The dominance of *Haha* (59.6%) in this cluster illustrates normalization and public satire of gender relations, though ridicule directed at feminist arguments is also a plausible reading.
- **Cultural Identity and Celebration:** content related to symbolic or national events such as ,_ l-Jl ol_l i, [International Women's Day] and national achievements (*e.g.,* ·-J·Jl ol_ll [Tunisian woman], ,,_ l ol_l [Egyptian woman]). Unlike "Gender Roles and Social Norms," which centers on contested interpersonal dynamics, this cluster is anchored in collective and institutional recognition of women's achievements and national pride. Engagement was overwhelmingly positive, reflecting collective identification with empowerment themes.

Taken together, these patterns indicate that empowerment discourse on Arabic Facebook is primarily affirmed (*Love*) or processed through humor (*Haha*), while *Angry*, *Sad*, and *Care* surface selectively, peaking for religion-inflected debates (*Angry*) and violence-related content (*Sad*/*Care*).

### 5.3. Key Findings

This exploratory analysis demonstrates how computational methods can illuminate the emotional

| Topic | Love | Haha | Wow | Sad | Angry | Care | DR | Posts |
|---|---|---|---|---|---|---|---|---|
| Violence and Discrimination | 41.6% | 37.1% | 3.1% | 7.8% | 5.8% | 4.7% | Love | 79,053 |
| Religion and Morality | 43.1% | 36.2% | 1.1% | 4.7% | 10.4% | 4.4% | Love | 13,326 |
| Policy, Law, and Institutions | 45.3% | 36.5% | 3.0% | 4.5% | 6.2% | 4.5% | Love | 8,750 |
| Gender Roles and Social Norms | 31.8% | 59.6% | 1.2% | 3.1% | 1.0% | 3.3% | Haha | 7,903 |
| Cultural Identity and Celebration | 50.5% | 38.6% | 0.8% | 4.4% | 1.3% | 4.3% | Love | 6,538 |

Table 2: Distribution of Facebook emotional reactions across the five dominant discourse topics related to women's empowerment. DR denotes the dominant reaction for each topic. Reaction percentages reflect aggregate counts across all posts in a topic cluster and should be interpreted as indicators of collective affective salience rather than individual stances.

and thematic structure of gender discourse in Arabic digital spaces. Key observations include:

- **Thematic diversity:** The dataset reveals up to twenty distinct but interpretable themes encompassing social, economic, and political dimensions of empowerment.
- **Emotion-topic alignment:** Topics emphasizing achievement, policy progress, and cultural recognition elicited predominantly positive emotions. Discussions addressing violence, discrimination, or injustice showed relatively higher shares of *Sad* and *Angry* reactions, indicating collective empathy and moral concern.
- **Reaction ambiguity:** Dominant reactions are best treated as collective affective indicators rather than unambiguous stance labels. Post-level annotation studies are needed to validate interpretations of individual reaction types.
- **Analytical transparency:** The frequency-based BERTopic framework allows direct traceability from clusters to lexical evidence, enhancing interpretability for social scientists.

Overall, this case study illustrates the potential of combining interpretable Arabic topic modeling with engagement analytics to uncover the interplay between thematic framing and emotional resonance in public discourse about women's empowerment.

# 6. Applications and Future Work

The presented corpus offers broad opportunities for advancing both computational research and social understanding of Arabic discourse on women's empowerment. Its combination of large-scale, multi-dialectal text, audience engagement signals, and contextual metadata positions it as a bridge between social science inquiry and natural language technology.

**Applications for Large Language Models and AI Safety** Beyond its immediate social research value, the dataset represents an important resource for the development and evaluation of LLMs in Arabic. It can support the fine-tuning of safety and moderation filters, enabling models to distinguish between empowerment discourse and harmful or abusive content. The inclusion of real-world audience reactions facilitates the creation of emotion-aware models that recognize empathy, pride, or anger within socio-political discussions. The corpus also offers a foundation for bias auditing and mitigation, allowing researchers to examine how generative models represent or respond to women's issues in Arabic contexts. Finally, the dataset can guide the development of contextually grounded and culturally sensitive LLMs capable of engaging constructively with sensitive gender-related topics.

**Computational Social Science Applications** The corpus opens multiple research avenues in computational social science and communication studies. It enables cross-cultural analyses of how women's empowerment is framed across different Arab countries, media ecosystems, and institutional contexts, revealing regional variations in discourse and audience reception. Its longitudinal structure supports the study of temporal dynamics, tracking how conversations around gender roles evolve over time in relation to major political, social, and cultural events. The inclusion of page-level metadata makes it possible to examine the diffusion of empowerment narratives across organizations, activists, and media outlets. The dataset's engagement metrics serve as a scalable proxy for public sentiment and social attitudes toward gender equality, providing a rich empirical foundation for understanding collective emotions in online discourse.

**Natural Language Processing and Machine Learning Research** For the NLP community, this corpus offers a rare and richly annotated foundation for advancing Arabic language technologies. Its multi-dialectal composition supports the development of models capable of handling linguistic variation across diverse Arabic dialects while predicting fine-grained emotional and affective categories. The thematic breadth of the data further enables interpretable and generalizable classifiers for social issue detection and topic categorization in Arabic discourse. By linking textual content to structured engagement signals, the corpus facilitates multimodal and context-aware modeling of language, emotion, and audience behavior.

**Digital Humanities and Cultural Studies** From a digital humanities perspective, the corpus serves as a large-scale window into Arab cultural discourse on gender and power. Scholars can examine recurring rhetorical patterns in empowerment narratives, such as representations of success, struggle, and resistance. Temporal comparisons allow for tracing shifts in how gender roles are articulated and contested in online spaces over the past decade. The geographical diversity of sources permits the comparison of engagement dynamics across Arab regions, revealing localized forms of digital activism and cultural reception.

**Future Directions** Future work will extend this corpus in several directions. First, we plan to develop benchmark tasks, including at least one classification task (e.g., dominant reaction prediction from post text) with baseline models, to characterize dataset difficulty and support community evaluation. Second, we plan to incorporate visual and audio data for richer multimodal analysis and to extend coverage to other social platforms such as X (formerly Twitter) and Instagram. Third, we will create benchmark tasks for Arabic emotion recognition, bias detection, and cultural safety evaluation. These efforts will further position the dataset as a cornerstone for both computational social science and responsible AI research in Arabic contexts.

## 7. Conclusion

The Arabic Women and Society Corpus offers a new resource for studying women's empowerment, social reform, and public discourse in Arabic digital spaces. With more than 252K posts from 77 countries and detailed audience engagement information, it provides the means to explore linguistic, cultural, and emotional patterns over a full decade. The dataset also contributes to the growing integration between computational linguistics and social research by enabling cross-disciplinary studies on digital participation and social change. The corpus, together with documentation and processing code, will be released for research purposes under a controlled-access agreement to support transparent and reproducible work in Arabic natural language processing and digital humanities.

## 8. Limitations

**Single Platform Focus:** The corpus is limited to Facebook data, which may not represent discourse patterns on other platforms.

**Public Content Bias:** Only public posts are included, potentially missing more private or intimate discussions about women's issues that occur in closed groups or personal profiles.

**CrowdTangle Coverage:** The dataset reflects CrowdTangle's coverage limitations, which may have affected the representation of smaller or newer pages.

**Dialectal Representation:** While the corpus includes multiple dialects, the distribution may not perfectly reflect population proportions or linguistic diversity across the Arab world. Dialect-level analysis will require the application of specialized dialect identification tools.

**Reaction Interpretability:** Facebook reactions are inherently ambiguous. A single *Angry* click may express outrage at injustice or disapproval of an empowerment message; a *Haha* click may reflect humor or ridicule. Reaction distributions should be treated as aggregate affective signals rather than unambiguous stance labels. Post-level annotation studies are needed to validate interpretations.

**Cultural Sensitivity:** Discussions of women's issues in Arab contexts involve complex cultural, religious, and political sensitivities that require careful interpretation by researchers with relevant domain expertise.

**Platform Evolution:** Facebook's algorithms, interface changes, and policy modifications over the decade may have influenced content visibility and engagement patterns.

**Digital Divide:** The corpus may underrepresent perspectives from communities with limited internet access or lower social media adoption rates.

**Event Bias:** Engagement patterns may be skewed by major events that generated exceptional social media activity during the collection period.

## 9. Ethical Considerations

**Privacy and Consent:** All content was collected from publicly available Facebook pages and groups via the CrowdTangle API prior to its discontinuation in August 2024. We accessed only administrator-designated "public" material; no personal profiles, private messages, or restricted content were included. To protect individual users, we applied a three-step anonymization procedure: (1) removal of all named user identifiers appearing in post metadata (e.g., user names and profile links); (2) replacement of personal names appearing within post text using a named entity recognition (NER) pass; and (3) removal of phone numbers, email addresses, and URLs linking to individual profiles. We acknowledge that "public" does not equal "consented," and that contextual integrity may be affected when repurposing platform data for research. Residual re-identification risk exists, particularly for high-profile public figures whose writing style or post content may be recognizable; researchers using the corpus should weigh this risk against the societal value of their study.

**Legal Basis and Licensing:** Data were collected via the CrowdTangle API under Meta's then-current terms of service for academic research. The corpus is distributed under a Creative Commons Attribution NonCommercial ShareAlike 4.0 (CC BY-NC-SA 4.0) License to the extent that applicable platform terms permit. Researchers are responsible for independently verifying that their intended use complies with Meta's current terms of service and any applicable national data protection legislation (including GDPR for data subjects in the EU).

**Institutional Review:** This study was reviewed by the Institutional Review Board (IRB) and was determined to be exempt from full IRB review under the applicable human-subjects research guidelines. The research follows institutional human-subjects policies and data minimization principles.

**Representation and Bias:** Social media discourse is not a population-representative sample. Participation, visibility, and platform curation can privilege some voices and marginalize others (e.g., users in conflict zones or under censorship). Models trained on such data may reproduce or amplify these biases; results should not be overgeneralized or deployed for high-stakes decisions without auditing and domain expertise.

**Cultural Competency:** Women's empowerment intersects sensitive domains (e.g., violence, legal rights, health, economic autonomy) that vary across Arab societies. Analyses should involve Arabic-speaking researchers and cultural experts to interpret topics and engagement patterns without imposing external frameworks or reinforcing stereotypes.

**Harm Prevention and Dual-Use:** Techniques for topic discovery and engagement profiling could be misused (e.g., surveillance, targeted harassment, manipulation). We discourage applications that identify individuals or optimize divisive messaging. Any downstream use should include risk assessment, mitigation plans, and human oversight.

**Data Collection and Context:** We aggregate engagement at the post level and do not track individual users across posts or infer personal attributes. This reduces privacy risks compared to user-level modeling, but residual risks remain (e.g., recontextualization). Researchers should weigh societal benefits against potential harms when designing studies or releasing artifacts.

**Stakeholder Impact and Accountability:** Findings may affect advocates, NGOs, platforms, and policymakers. We recommend participatory design where feasible, clear governance for any operational use (e.g., moderation workflows), appeal mechanisms, external audits, and public reporting of performance across groups.

**Data Sharing and Access Controls:** To reduce re-identification and misuse risk, the full dataset is not publicly released without restriction. Access for non-commercial academic use is granted via a data-sharing agreement requiring (a) institutional ethics or IRB approval or equivalent waiver, (b) scope limits aligned with the stated research purpose, and (c) a prohibition on de-anonymization. We provide methodological details (models, parameters, evaluation) to support replication on alternative data. The data-sharing request form is available online[4].

**Institutional Review and Compliance:** This work is designed to comply with applicable regulations (e.g., GDPR for EU data). Researchers using the corpus remain responsible for local legal and ethical compliance in their jurisdictions.

## Acknowledgments

This work was made possible by the National Priorities Research Program (NPRP) grant NPRP14C-0916-210015 from the Qatar National Research Fund (QNRF), a member of the Qatar Research, Development and Innovation Council (QRDI).

## References

[4]https://tinyurl.com/4ke5jwyw